# From LIMA to DeepLIMA: following a new path of interoperability

Victor Bocharov[1], Romaric Besançon[1], Gaël de Chalendar[1*], Olivier Ferret[1] and Nasredine Semmar[1]

[1*]Université Paris-Saclay, CEA, List, Palaiseau, France.

*Corresponding author(s). E-mail(s): gael.de-chalendar@cea.fr;
Contributing authors: victor.bocharov@cea.fr;
romaric.besancon@cea.fr; olivier.ferret@cea.fr;
nasredine.semmar@cea.fr;

## Abstract

In this article, we describe the architecture of the LIMA (Libre Multilingual Analyzer) framework and its recent evolution with the addition of new text analysis modules based on deep neural networks. We extended the functionality of LIMA in terms of the number of supported languages while preserving existing configurable architecture and the availability of previously developed rule-based and statistical analysis components. Models were trained for more than 60 languages on the Universal Dependencies 2.5 corpora, WikiNer corpora, and CoNLL-03 dataset. Universal Dependencies allowed us to increase the number of supported languages and to generate models that could be integrated into other platforms. This integration of ubiquitous Deep Learning Natural Language Processing models and the use of standard annotated collections using Universal Dependencies can be viewed as a new path of interoperability, through the normalization of models and data, that are complementary to a more standard technical interoperability, implemented in LIMA through services available in Docker containers on Docker Hub.

# Declarations

*Type of submission*

Project Notes

*Funding*

Internal. Partly supported by several French national and European projects.

*Conflicts of interest/Competing interests*

None.

*Availability of data and material*

All data used in this work are open-source.

*Code availability*

All the source code of the software described in this work is open-source, released under a free software license.

# 1 Introduction

The history of Language Technologies since the mid-1990s is closely linked to a series of open-source tools and platforms, starting with GATE (Cunningham et al, 2002b) and currently continuing with the recent Deep Learning approaches, implemented in linguistic analyzers such as spaCy or Stanza (Qi et al, 2020). While some of them, such as OpenNLP, are still attached to the context in which they were created, others, like Stanford CoreNLP (Manning et al, 2014), have evolved to integrate new paradigms. In this article, we present the LIMA linguistic analysis platform and more precisely the work we have done for performing such evolution by integrating into LIMA, initially built on rule-based components, new components based on recent Deep Learning techniques.

The LIMA analyzer offers a wide range of text analysis methods available through a unified interface in its configurable general-purpose text analysis framework. It is currently used in research and commercial projects. The framework provides abstract concepts of the analysis as a sequence of steps, shared linguistic resources, and an internal representation of analysis data. Specific text processing methods are wrapped into modules that can be used together and form an analysis pipeline. Recently, we added several text processing units based on modern deep neural networks. These new processing units implement the same interfaces and are interoperable with existing ones. This way, mixed pipelines including both rule-based and neural network-based processing steps are possible.

This evolution of the LIMA analyzer can be viewed as an adaptation to a recent *de facto* standard for linguistic analyzers resulting from the successful development of machine learning and deep learning approaches. The most recent linguistic analyzers such as spaCy or Stanza are based on deep learning models, which require significantly large annotated corpora for their training. These corpora are obviously the bottleneck of this paradigm since the cost of their production is very high and as such, they lead to a particular kind of interoperability: two linguistic analyzers can be based on different models but they are trained on the same corpora. This dependency has in practice a significant impact on the analyzers and tends to normalize several of their dimensions, such as their input format, their tokenization, their part of speech tagset, or the kind of syntactic analysis they perform. While this dependency can be viewed as a restriction, the Universal Dependencies (UD) project (Nivre et al, 2016) has illustrated its power, with the possibility for an analyzer such as LIMA to process six times more languages than before by exploiting the UD annotated corpora with only a limited effort. More globally, this new paradigm has also led to the normalization of other aspects of linguistic analyzers. For instance, since they are based on deep learning models, they also adopted standard frameworks for this kind of models, Tensorflow (Abadi et al, 2015) and PyTorch (Paszke et al, 2019), and the same programming language, Python. While we can consider that some aspects of interoperability resulting from this new paradigm are taken into account by existing work about interoperability (Rehm et al, 2020), such as the necessity to share a form of semantic space, under the form of annotations in the considered case, the kind of interoperability emerging from this convergence is probably more difficult to define and does not necessarily fit well the existing analysis of this notion.

In the remainder of this paper, we briefly mention several existing widely known text analysis frameworks and tools. We then describe the design of the LIMA framework and the way the analysis is represented internally. Finally, we provide details about the integration into LIMA of processing units based on deep neural networks and their exploitation on the full set of UD languages.

## 2 Related work

### 2.1 Natural Language Processing frameworks and toolkits

The search for Natural Language Processing (NLP) toolkits or frameworks on the Web leads to confusing results with very heterogeneous lists of tools including Deep Learning frameworks (PyTorch, Tensorflow...) that can be used for implementing NLP models, word embedding models, or various linguistic tools. In this article, we clearly focus on linguistic analyzers as they were defined before the Deep Learning wave, which already covers a significantly large number of recent and older tools that we quickly review hereafter:

*GATE*

(General Architecture for Text Engineering) (Cunningham et al, 2011) is an open-source software toolkit originally developed at the University of Sheffield in 1995. GATE includes many analysis modules (processing resources), a graphical environment, and an information extraction system called ANNIE (A Nearly-New Information Extraction System) (Cunningham et al, 2002a).

### UIMA

(Unstructured Information Management Architecture) (Ferrucci et al, 2009) is an OASIS[1] standard for content analysis developed at IBM, and Apache UIMA is an open-source implementation of this standard. DKPro (The Darmstadt Knowledge Processing Software Repository) (Eckart de Castilho and Gurevych, 2014) is a collection of software components for NLP based on the Apache UIMA framework.

Both GATE and UIMA provide pipeline-based frameworks and analysis modules. Within the GATE universe, modules are mostly Java-developed. Apache UIMA includes both Java and C++ frameworks and annotation modules can be written in Java, C++, Perl, Python, and TCL.

### Apache OpenNLP[2]

is a machine learning library that provides analysis components for many NLP tasks: language detection, tokenization, part of speech tagging, named entity recognition (NER), parsing, and coreference resolution. This is also a Java-based toolkit.

### NLTK

(Natural Language Toolkit)[3] (Bird et al, 2009) is a set of Python libraries for solving natural language processing tasks. In addition to analysis modules, NLTK also includes corpora and lexical resources available through the same installer.

### Stanford CoreNLP

is a Java annotation pipeline framework, which provides different NLP steps such as tokenization, sentence splitting, morphological analysis, part of speech tagging, Named Entities Recognition (NER), syntactic parsing, and coreference resolution. Stanford CoreNLP is available in several languages like Chinese and English. Support for other languages is less complete, but many of the NLP tools also support models for French, German, and Arabic, and building models for further languages is possible. *Stanza* is the new NLP pipeline by the Stanford NLP group implementing UD parsing for more than 70 languages, named entity recognition, and sentiment analysis[4]. Stanza is based on neural models implemented in PyTorch. It includes a Python binding to CoreNLP.

### spaCy[5]

is another open-source Python library offering software components for text analysis. spaCy is partially implemented using Cython[6] and their authors claim that their main focus is to provide an industrial instrument that is capable of operating at a large scale. AllenNLP[7] is a framework for deep Learning workflows created on top of spaCy and the PyTorch machine learning library.

---

[1] https://www.oasis-open.org/
[2] http://opennlp.apache.org/
[3] https://www.nltk.org/
[4] https://stanfordnlp.github.io/stanza/index.html
[5] https://spacy.io/
[6] A compiled language that offers better performance and memory management for Python-like code.
[7] https://allennlp.org/

*UDPipe*

(Straka and Strakova, 2017) is an open-source tool that implements NLP tasks required to reproduce Universal Dependencies 2.0 annotations: tokenization, sentence segmentation, part of speech tagging, lemmatization, and dependency parsing. UDPipe provides both training and annotation functionalities. The training part uses only Universal Dependencies annotation without any supplementary data. UDPipe is written in C++ and bindings for Python, Perl, Java, and C# are provided.

*UDify*

(Kondratyuk and Straka, 2019) is based on a single model for the analysis of 75 languages with a BERT-based encoder. It uses cased BERT-Base multilingual model pre-trained on Wikipedia dumps for 104 languages[8]. The original research paper describes different fine-tuning strategies and their effect on high-resource and low-resource languages.

Apart from how old they are, these various tools can be categorized according to different criteria. For instance, GATE and UIMA are first defined as architectures, even if they also cover a large set of linguistic modules implementing their architecture requirements. Conversely, NLTK wraps standard existing tools for using them in Python but without adopting strong unifying principles. Apache OpenNLP, UDPipe, UDify, and Stanza are all defined through fairly homogeneous machine learning models, statistical models for Apache OpenNLP and Deep Learning models for the others while Stanford CoreNLP and spaCy are more heterogeneous in terms of types of components, mixing machine learning and rule-based modules. According to this coarse-grained taxonomy, LIMA is globally close to tools such as Stanford CoreNLP and spaCy but is based on a set of design principles that links it to some extent to architectures such as GATE and UIMA.

## 2.2 Universal Dependencies

Universal Dependencies (UD) (Nivre et al, 2016) is an international project and a multilingual annotation framework that provides a universal inventory of linguistic categories and annotation guidelines covering tokenization, part of speech and feature tagging, and dependency parsing. Within the UD project, a cross-linguistically consistent treebank annotation for many languages is created. A new version of the Universal Dependencies treebank collection is released twice a year. Current version UD 2.8 includes 202 treebanks for 114 languages.

There exists a wide range of software tools[9] (editors, visualizing tools, consistency checkers, and libraries) that work with Universal Dependencies annotations.

# 3 Design of the LIMA framework

LIMA is a C++ toolkit and a pipeline-based analysis framework developed by the LASTI laboratory of CEA List. It was designed and developed with several objectives

---

[8] https://github.com/google-research/bert/blob/master/multilingual.md
[9] https://universaldependencies.org/tools.html

that encompass the features of all the linguistic analysis tools and frameworks listed above:

- multilingualism: an ability to work with a broad spectrum of languages;
- diversity of use cases: LIMA must be useful as a basic component for various text processing applications such as information retrieval, automatic summarization, machine translation, etc.;
- extensibility: its architecture makes it possible to easily add new functionalities or replace the implementation of existing components;
- efficiency: LIMA must be able to process large corpora and to work in an industrial context.

### 3.1 Pipeline

Many applications performing different types of linguistic analysis share some common features: the pipeline-based design that implies sequential execution of relatively independent actions is one of them. The core of LIMA is a framework that implements an abstract text processing pipeline, which includes the following concepts:

- analysis data: results of the linguistic analyses consisting of a set of named layers, each of them containing a specific type of information (tokens, parts of speech, syntactic relations...);
- processing units: software modules creating or modifying the analysis data;
- linguistic resources: language-specific data that are used by processing units. They can be shared by several of them.

The processing units implement some analysis methods and are generally language-agnostic. All language-specific data are represented in the form of linguistic resources. Figure 1 gives a high-level overview of the design of the LIMA framework.

A pipeline is defined in a LIMA configuration file as a sequence of the processing steps that will be executed unconditionally for any input data. Each processing step is defined by either a nested list of other processing steps or a single processing unit associated with its configuration parameters. The framework doesn't specify any particular behavior or roles for processing units except that they will be executed sequentially in the given order. Processing units are not aware of the particular list of other processing units executed before or after but are defined by the analysis layers expected as input in the analysis data or created as output.

For instance, a typical rule-based pipeline for NER includes a tokenization step that starts from raw text and does tokenization and sentence segmentation, a part of speech tagging step, one or several sequentially executed NER tagging steps, and a dumping step that generates the output.

### 3.2 Analysis data

Analysis data in LIMA is a set of named layers of arbitrary type. There is no particular restriction on the possible content of each layer: it depends only on the capabilities of the processing units to interpret it. However, many units treat the text being analyzed as a directed graph with tokens as nodes and edges representing the sequence of tokens in the text. Annotations of arbitrary types can be attached to graph nodes. The graph can contain multiple paths with different nodes representing alternative versions of tokenization or alternative interpretations (an example of such a graph is shown in Figure 2).

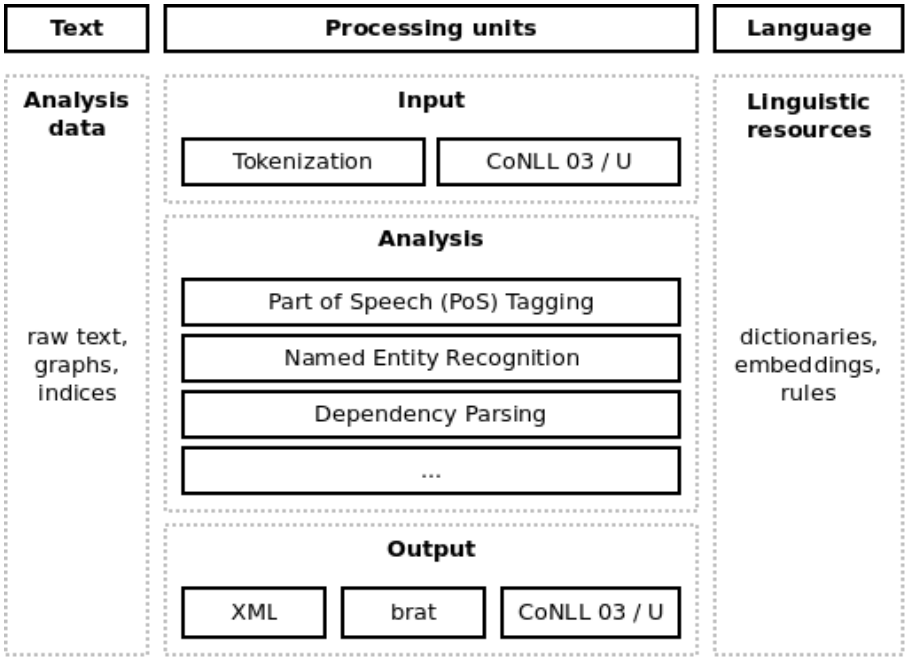

**Fig. 1** LIMA framework design

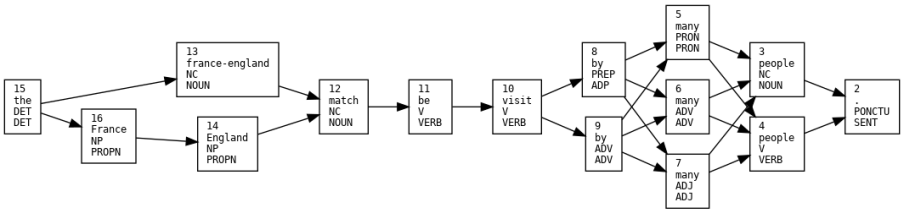

**Fig. 2** The analysis graph with ambiguities on both tokenization and part of speech tagging levels for the sentence "The France-England match is visited by many people.".

Processing units are able to edit graphs by adding, deleting, or replacing nodes and edges and modifying attached interpretations. The analysis can contain multiple layers with graph-based text representations. Additional layers representing links between these graphs can be added if necessary.

### 3.3 Preconfigured pipelines in LIMA

Table 1 summarizes the pipelines in LIMA that are related to NER and UD parsing tasks and are considered as "production-ready".

| Name | Task | Languages |
| --- | --- | --- |
| deepud | UD parsing | All UD languages |
| deepud-pretok | UD parsing | All UD languages |
| ner-rules | NER | English, French |
| ner-rules-pretok | NER | English, French |
| ner-deep | NER | English, French |
| ner-deep-pretok | NER | English, French |

**Table 1** "Production-ready" pipelines in LIMA: "pretok" pipelines expect pre-tokenized input in CoNLL-U form while others expect raw text.

"deepud" and "deepud-pretok" pipelines use Deep Neural Networks (DNNs) for all processing stages. NER pipelines use RNN for tokenization and either a rule-based NER module or one based on deep learning techniques.

## 4 Processing units based on deep neural networks

The rule-based implementations of processing units exist for a long time in LIMA (Besançon et al, 2010). They offer high flexibility in configuration and explainability of analysis results. However, the development of these rule-based modules is expensive as it involves a lot of manual effort. Moreover, most of the rule-based analysis components are difficult to port from one language or domain to another. Machine learning methods are easier in portability and with recent progress in deep neural network architectures, they outperform many rule-based methods in analysis quality. Below, we describe our DNN-based modules.

### 4.1 Named entity recognition

There are two implementations of NER processing units in LIMA: the first one based on hand-crafted rules (used in ner-rules pipelines) and the other one DNN-based trained on CoNLL-03 dataset for English and WikiNER for French (used in ner-deep pipelines). They are different in the set of entities they extract (see Table 2), in their processing steps, and in their performance.

The analysis performance between these two approaches can be compared only for common entity types. In Table 3, we provide performance figures for English on both WikiNer and CoNLL-03 corpora.

DNN-based NER for English has been trained on CoNLL-03 data. The rule-based NER has been developed with CoNLL-03 in mind. This explains better results on the CoNLL-03 corpus for both methods. At the same time, *ner-deep* significantly outperforms *ner-rules* in all cases.

The processing speed of rule-based NER depends on the number of rules applied (see Table 4). In either case, the rule-based NER is much faster than the DNN-based one. It is also easier to develop a limited set of dedicated rules for a new set of entities in a new application domain than to create the annotated corpus necessary to train DNN-based models. Thus the flexible and easy-to-use rule-based NER remains a very important feature of LIMA. We also often use both approaches together: rules are first manually written to extract a large part of the entities that are simple to describe. These rules are applied to a corpus from the target domain. The resulting annotations are then manually corrected on a little part of the corpus. This corrected segment is used to train a neural model which is applied to a new part of the corpus. This extended training corpus is used to train a new version of the model, and so on. This iterative process allows us to quickly build efficient NER models on new domains.

| Entity type | ner-rules | ner-deep |
|---|---|---|
| Number | + | - |
| Date and time | + | - |
| Organization | + | + |
| Location | + | + |
| Person | + | + |
| Event | + | - |
| Product | + | - |
| Miscellaneous | + | + |

**Table 2** The types of entities the NER pipelines can recognize.

| Entity type | ner-rules | | ner-deep | |
|---|---|---|---|---|
| | WikiNer | CoNLL-03 | WikiNer | CoNLL-03 |
| Organization | 38.95 | 29.07 | 51.02 | 86.02 |
| Location | 59.98 | 73.10 | 72.14 | 91.51 |
| Person | 64.66 | 74.77 | 74.34 | 95.27 |
| Miscellaneous | 3.63 | 7.65 | 42.05 | 76.88 |
| All entities | 50.43 | 59.15 | 63.04 | 89.13 |

**Table 3** NER performance: F1 score for English on two datasets.

| Language | ner-rules | ner-deep |
|---|---|---|
| English | 6,433 | 295 |
| French | 12,962 | 272 |

**Table 4** NER speed (tokens per second).

## 4.2 Universal Dependencies parsing

"deepud" and "deepud-pretok" pipelines implement part of speech tagging and dependencies parsing trained on corpora from UD collection. Figure 3 shows the analysis graph corresponding to the possible results of the processing with "deepud" pipeline. The following processing steps are performed: (only "deepud") tokenization and sentence segmentation; (only "deepud-pretok") pre-tokenized text reading from CoNLL-U formatted file; part of speech and feature tagging and dependency parsing (one processing unit does both); lemmatization; output in CoNLL-U or another format (see Figure 4).

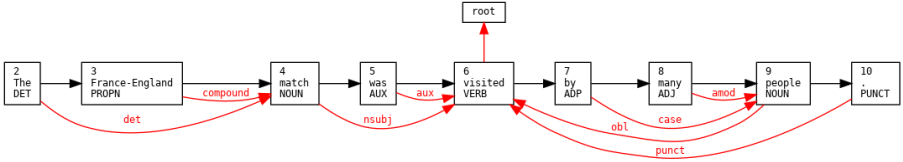

**Fig. 3** The analysis graph after the UD parsing: black edges: sequence; red edges: syntactic dependencies.

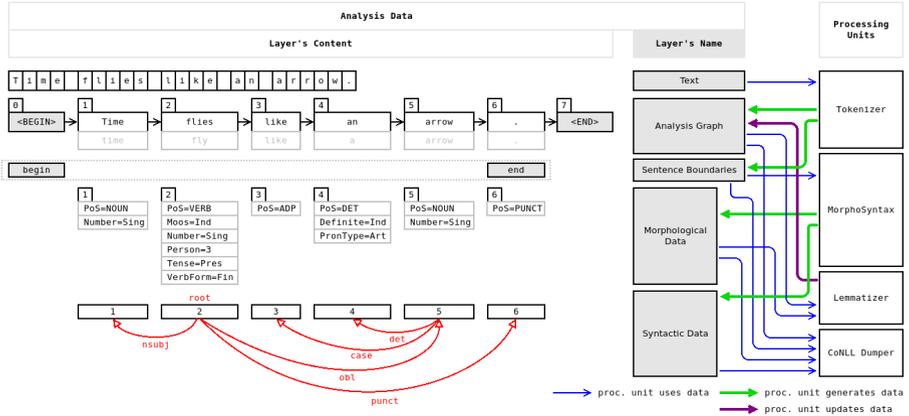

**Fig. 4** UD parsing pipeline (the right side) and the corresponding analysis data (the left side).

On average for the LAS score, LIMA nearly equals or outperforms UDPipe and UDify (see Table 5). The list of possible morphological features in LIMA is shorter than in the corresponding treebanks: rare features are omitted. This explains the score on features, lower than the UDPipe one. A performance comparison between LIMA and UDPipe and UDify, for all supported languages, is presented in Figure 5.[10]. These results show that LIMA outperforms UDPipe for most languages, but UDify often obtains better results, except for a set of low-resource languages (on the left of the figure).

| **Tool** | UPOS | UFeats | Lemmas | UAS | LAS | Speed |
|---|---|---|---|---|---|---|
| LIMA | .92 | .81 | .89 | .82 | .75 | 349 |
| UDPipe | .94 | .90 | .94 | .81 | .75 | 2761 |
| UDify | .89 | .81 | .87 | .82 | .74 | 91 |

**Table 5** UD parsing performance averaged on all UD languages (speed in tokens per second).

In the subsections below, we give details on each of the deep learning-based pipeline units.

---

[10] The exhaustive score comparison can be found on LIMA's GitHub page: https://github.com/aymara/lima-models/blob/master/eval.md

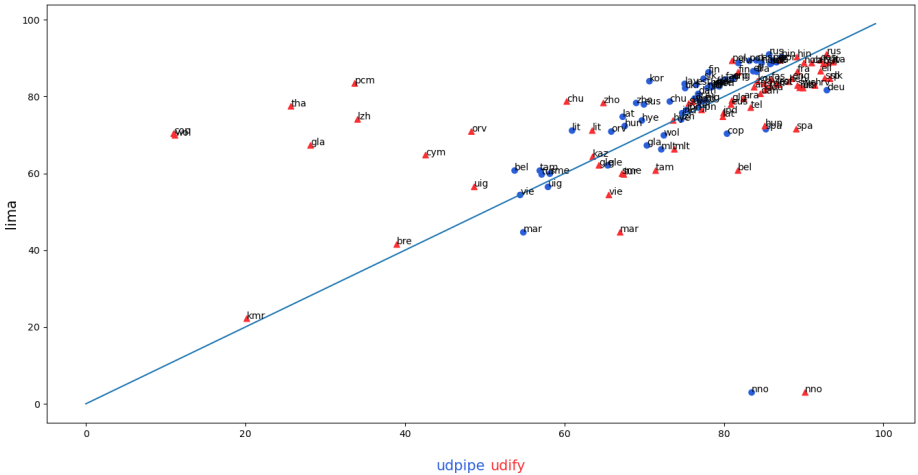

**Fig. 5** Comparative LAS scores between LIMA and UDPipe/UDify: each dot corresponds to a language, y-axis marks the score with LIMA, x-axis marks the score for UDPipe (resp. UDify) for the blue dots (resp. red triangles). The points above the diagonal therefore correspond to better results with LIMA.

### 4.2.1 Tokenizer

For token and sentence segmentation, we adapted the character labeling approach proposed in Universal segmenter (Shao et al, 2018). It is based on bidirectional recurrent neural networks with conditional random fields (BiRNN-CRF) and Viterbi decoder. The output tagset consists of token segmentation tags and sentence segmentation tags.

Three concatenated embeddings are given on each RNN step: the unigram embedding (current character only), the embedding of the bigram including previous character and the current one, and the embedding of the trigram that includes previous, current, and next characters.

The dimension of each embedding is computed at training time and depends on the number of different n-grams of a given length found in the training set. This makes the model smaller for most languages without significantly reducing its quality.

### 4.2.2 Morphological tagger and dependency parser

The morphological tagger assigns part of speech tags and feature tags for each word in the sentence. For this purpose, we use a similar sequence labeling approach as described above for tokenization. As soon as there are many different types of tags (part of speech tags, number, gender, case, etc.), a dedicated classifier is required for each type. We use single BiRNN input for all types of tags with different CRF outputs for each classifier. CRF outputs for taggers are connected to a second BiRNN layer. The remaining layers are used by the dependency parser only.

The part of speech tagging performance of LIMA is good but the feature annotation must be improved, lying around 10 points behind the compared systems.

For dependency parsing, we adapted a graph-based parser (Kiperwasser and Goldberg, 2016) with the deep biaffine attention arc scoring method of (Dozat and Manning, 2017). Arc scorer is attached on top of the concatenation of the output of the same BiRNN that is used for morphological tagging and dedicated BiRNN that is used for dependency parsing only. All these tasks (i.e. morphological tagging and dependency parsing) are trained simultaneously.

The input of the BiRNN shared by the taggers and the parser consists of pre-trained word embeddings for all words and trainable word embeddings for frequent words. The sum of word embeddings of these two types is concatenated with the final state of character-level RNN for the corresponding word.

FastText word embeddings with subword information are used as pre-trained word embeddings (Bojanowski et al, 2016). The choice of fastText instead of word2vec (Mikolov et al, 2013) or Glove (Pennington et al, 2014) is made for two reasons: fastText provides pre-trained models (Grave et al, 2018) for most languages available in the Universal Dependencies collection, and subword information gives meaningful word vectors, even for Out-Of-Vocabulary words.

### 4.2.3 Lemmatizer

The lemmatizer uses the source form of the word and morphological tags (part of speech and features tags) predicted in the previous step. The lemmatization task is handled as a sequence-to-sequence translation problem at the character level. Our approach is similar to the one adopted in Turku Neural Parser Pipeline (Kanerva et al, 2018): surface word form together with predicted tags is given as input. We add morphological information in the form of embeddings of tags to the encoder at the initial state.

Lemmatization as it is described above is a context-independent task: the neighboring words aren't used to predict lemma.

## 5 Technical details

Deep neural networks in LIMA are implemented using the TensorFlow framework. The training code is written in Python[11] and models are loaded in the C++ processing units for inference. The Python interface to TensorFlow can be easily installed on most modern operating systems via corresponding package managers (pip, conda) but this is not the case for its C interface. For deployment, we package TensorFlow shared library in the form of a Debian package and we distribute it via launchpad. The installer for Windows is also provided.

The typical size of a fastText embedding file for one language is about 7Gb. To reduce both the memory consumption during inference and the network traffic we deploy quantized fastText embeddings instead of original ones (Bocharov, 2020). The quantization with an eight-time size reduction doesn't notably affect the analysis performance (Bocharov and de Chalendar, 2020). We can thus use embeddings with a size around 600MB for each language instead of the original 7GB.

---

[11] https://github.com/aymara/{lima-ud-segmenter,lima-ud-depparser,lima-ud-lemmatizer,lima-tfner}

## 6 Platforms, interoperability, and LIMA

LIMA has always been developed with the objective of being easily integrated into applications that need NLP. Its plugin-based architecture with dumpers and handlers makes it extremely generic. Initially designed for integration into a distributed search engine architecture but with no prior notions about the distribution method, LIMA can easily be adapted to be integrated into many platforms. It has, for instance, already been successfully integrated into a UIMA architecture. At the architectural level, LIMA has interesting interoperability capabilities.

A second aspect concerning interoperability is that of data and annotations. The adoption of Universal Dependencies linguistic framework has the great practical advantage of unifying linguistic annotations for all languages, at the cost of a difficulty in representing some linguistic phenomena (Osborne and Gerdes, 2019) which is not a problem at the practical level for most applications that use LIMA.

Finally, by its use of a double licensing model, proprietary when in the need of advanced support and/or secrecy; and free software (currently AGPL and soon MIT or BSD), and thanks to its features described above, LIMA is able to efficiently participate to the eight objectives of the European Language Grid initiative[12]: business, societal, community, and technology related.

## 7 Conclusion and future work

In this article, we present the recent developments of LIMA, a linguistic analyzer whose very flexible architecture offers great adaptability for dealing with new languages or new domains without penalizing its performance in terms of speed. More precisely, we show how this flexibility can be used for adapting LIMA to a new implicit standard for linguistic analyzers based on Deep Learning models and trained on the resources of the Universal Dependencies framework (tags and annotated corpora). This adaptation led to the integration into LIMA of POS taggers and syntactic parsers for more than 60 languages with a level of performance comparable or higher to that of reference analyzers such as UDPipe or UDify and a good capability for processing low-resource languages. Moreover, we have shown that thanks to its flexibility, LIMA is able to easily combine Deep Learning and rule-based modules in the same architecture, which enables the most appropriate approach to be used for a given task.

There are several directions of future work to further improve LIMA. We will increase the number of supported languages with low-resourced ones where the training set is not available yet. We plan to improve analysis performance by employing techniques such as the distillation of transformer-based models. We will also improve inference speed with caching techniques and replace still large fastText embeddings with even smaller ones to reduce memory consumption. Finally, we will replace our dependence on TensorFlow with a deep learning framework that is more readily available in most Linux distributions as well as in MS Windows and MacOS.

---

[12] https://www.european-language-grid.eu/objectives/

We always try to make the use of LIMA easier on all platforms. The current step towards this goal is the improvement of its Python bindings and its Docker container.

## Acknowledgments

This publication was made possible by the use of the FactoryIA supercomputer, financially supported by the Ile-de-France Regional Council.